\documentclass[runningheads]{llncs}
\usepackage{graphicx}
\begin{document}
\title{Personalized Step Counting Using Wearable Sensors: A Domain Adapted LSTM Network Approach} 
%
%
\author{Arvind Pillai\and
Halsey Lea \and
Faisal Khan \and
Glynn Dennis
}
\authorrunning{A. Pillai et al.}
\titlerunning{Personalized Step Counting with an LSTM Network}
%
\institute{Artificial Intelligence \& Analytics, Data Science \& Artificial Intelligence, R\&D, AstraZeneca, Gaithersburg, Maryland, USA \\
\email{glynn.dennis@astrazeneca.com}}
\maketitle              
\begin{abstract}
Activity monitors are widely used to measure various physical activities (PA) as an indicator of general health. Similarly, real-time monitoring of longitudinal trends in step count has significant clinical potential as a personalized measure of disease related changes in daily activity.  However, inconsistent step count accuracy across vendors, body locations and individual gait differences limits clinical decision making. The tri-axial accelerometer inside PA monitors can be exploited to improve step count accuracy across devices and individuals. In this study, we hypothesize: (1) raw tri-axial sensor data can be modelled to create reliable and accurate step count, and (2) a generalized step count model can then be efficiently adapted to each person’s unique gait pattern using very little new data. Firstly, open-source raw sensor data was used to construct a long short term memory (LSTM) deep neural network to predict step count accurately ($\sim$98\%).  Then we generated a new, fully independent dataset using a different device and different subjects. Finally, a small amount of subject-specific data was domain adapted to produce personalized models with high individualized step count accuracy ($\sim$98\%). These results suggest models trained using large freely available datasets can be fine-tuned on patient populations lacking large historical data sets.  

\end{abstract}
\section{Introduction}
In 2008 the US government released its first-ever evidence-based guidance on PA, which acknowledged very strong evidence that physically active people have higher levels of health-related fitness, a lower risk profile for developing cardiovascular diseases, and lower rates of various chronic diseases than people who are not active \cite{tudor-locke}. A simple and inexpensive option for monitoring steps and cardiovascular health is the physical activity monitor. Step count is a quantifiable metric which demonstrates individuals taking $\geq$5,000 steps/day had substantially lower prevalence of adverse cardio-metabolic health indicators than those taking lower amounts \cite{schmidt}. In addition, step count has been shown to be inversely related to risk of progression to diabetes \cite{kraus}.  Collectively, reduced physical activity as measured by step count is a consistent predictor of death in chronic heart failure, possibly surpassing traditional laboratory-based exercise tests.

Recent advances in deep learning have enabled more sophisticated modelling of patterns in sensor data to classify human physical activities. LSTM based architectures have shown their efficacy in accurately predicting step count \cite{edel,chen}. Herein we describe a personalized step count algorithm that adapts to a subject's unique gait pattern with very little new subject-specific data collected from a different wearable device. We consider an LSTM deep learning algorithm to model human steps from two different wrist-worn devices containing accelerometers and gyroscopes. We observed a general LSTM model trained on publicly available data performs well for most subjects, but exhibited lower step count accuracy for some individuals.  We hypothesized that the observed decrease in step count accuracy may be due to individual subject gait differences and change in devices. Therefore, we show that an adaptive step count algorithm can be used to train personalized models to obtain accurate step count across devices and subjects.

\section{Materials and Methods}
In this paper, we use two datasets: (1) the "Pedometer" dataset \cite{ryan},
collected at Clemson University  using a Shimmer3 device from 30 subjects and (2) a fully independent, analogous "AZ" dataset, collected at AstraZeneca using an ActiGraph GT9X Link from 5 different subjects. The participants were asked to perform the 6-minute walk test in an indoor looped setting. Next, the steps and distance were manually annotated using video recordings from an Iphone XR. Morever, the rationale for developing this dataset was to test the ability of an LSTM step count algorithm to learn features that generalize across different device manufacturers and subjects. Accelerometer and gyroscope data from the wrist worn devices were used to model step count as a classification problem with left and right step as the two classes. The sensor data was resampled to 15Hz and windowed into $0.4\overline{6}$s readings based on tuning experiments and literature \cite{pachi}.

The deep learning model architecture consists of a many-to-one 256 unit LSTM followed by two fully connected layers of 512 and 256 units, respectively, with dropouts and activated using ReLU. Finally, a softmax layer was used to predict left or right step, and every transition between the two classes is counted as a step. The input features to the deep learning model architecture consists of individual windows with 3 readings from a tri-axial accelerometer(gs) and 3 from a tri-axial gyroscope(deg/sec). In order to personalize our model without the need to train from scratch, we investigated different variations of Transfer Learning. Domain adaptation is one such paradigm where the source and target probability distributions are different, i.e. given a source feature space $X_S$ and target feature space $X_T$, source label space $Y_S$ and target label space $Y_T$, we have $X_S=X_T$, $Y_S = Y_T$, and $P(X_S) \neq P(X_T)$ \cite{avi}.  In the context of our problem, $X_G=X_P^S$, $Y_G=Y_P^S$, and $P(X_G) \neq P(X_P^S)$, where \textit{G} represents a general model trained using Pedometer dataset (source) and P represents a personalized model for every subject \textit{S} in the AZ dataset (target).

The metrics chosen to evaluate our model are class accuracy and step count as shown in (1) and (2).
\begin{equation}
    accuracy_{class} = \left(\frac{N_R + N_L}{N_T}\right) * 100
\end{equation}
\begin{equation}
    accuracy_{steps} = \left(1 - \left|\frac{step\_count_{predicted}-step\_count_{ground truth}}{step\_count_{ground truth}}\right|\right) * 100
\end{equation}

where $N_R$ and $N_L$ are number of correctly classified right and left step windows respectively, and $N_T$ is the total number of steps. $step\_count_{ground truth}$ is obtain using visually confirmed annotation from the datasets and $step\_count_{predicted}$ is the step count predicted by the model.

\section{Results}
To establish a baseline, we obtained data from Pedometer dataset \cite{ryan} and implemented Piece-wise Aggregate Approximation (PAA) + merge approach described in \cite{feng}. 

After establishing a baseline, the LSTM deep learning model was designed as described in the previous section. There are three major experimental setups to assess the performance of our model. Firstly, a general model was trained and tested using only the Pedometer dataset (Shimmer3) using Leave-two-subject-out cross-validation (CV), we call this LSTM-General I. Next, this pre-trained general model was validated using the entire AZ dataset (ActiGraph GT9X Link) as a test set, which is documented as experiment LSTM-General II. Finally, a personalized model, LSTM-Personalized was domain adapted using a 2-step training process from the pre-trained general model using only 30 seconds of subject-specific data, and the test dataset was created by removing this information from each subject. As shown in Table 1, the median $accuracy_{class}$ and median $accuracy_{steps}$ among subjects were recorded for the three settings. Since, LSTM-General I generates different models for each CV fold, we calculate the mean of medians and use the best pre-trained model. 

The model was tuned based on window size, sensor modalities, device location, and number of units in each layer. In addition to training with different combination of layers, we also examined different LSTM architectures including stacked, bidirectional, and stateful.

\begin{table}
\caption{Summary of $\mathbf{accuracy_{class}}$ and $\mathbf{accuracy_{steps}}$ for all experimental setups.}\label{tab1}
\begin{tabular}{|l|l|l|l|l|}
\hline
{\bfseries Approach} &  {\bfseries Training Data} & {\bfseries Testing Data} & { $\mathbf{accuracy_{class}}$} & {$\mathbf{accuracy_{steps}}$}\\
\hline
PAA + Merge (Baseline) & N/A & Pedometer & N/A & 89.73 \\
PAA + Merge (Baseline) & N/A & AZ & N/A & 86.05 \\
LSTM-General I & Pedometer & Pedometer & 85.98 $\pm\,2.46$ & 98.63 $\pm\,2.37$\\
LSTM-General II & Pedometer & AZ & 60.01 & 94.49\\
LSTM-Personalized & Pedometer + $\sim$AZ & AZ & 91.76 & 98.81\\
\hline
\end{tabular}
\end{table}

As shown in Table 1, LSTM-General I improves $accuracy_{steps}$ by $\sim$10\% from the baseline, and the standard deviation indicates a left-skewed distribution. Moreover, the baseline tested is an approach that does not classify steps, so $accuracy_{class}$ metric is not applicable. Additionally, there is a decrease in performance of LSTM-General II which is indicated by a 25.97\% and 4.14\% drop in $accuracy_{class}$ and $accuracy_{steps}$ respectively, which shows lack of robustness in handling different devices. Next, we observe LSTM-Personalized outperforms LSTM-General II as indicated by a 31.75\% and 4.32\% increase in $accuracy_{class}$ and $accuracy_{steps}$ respectively. This implies only a small amount of labelled data (represented as $\sim$AZ in Table 1) is necessary to improve performance.

\begin{figure}
\centering
\includegraphics[width=6cm]{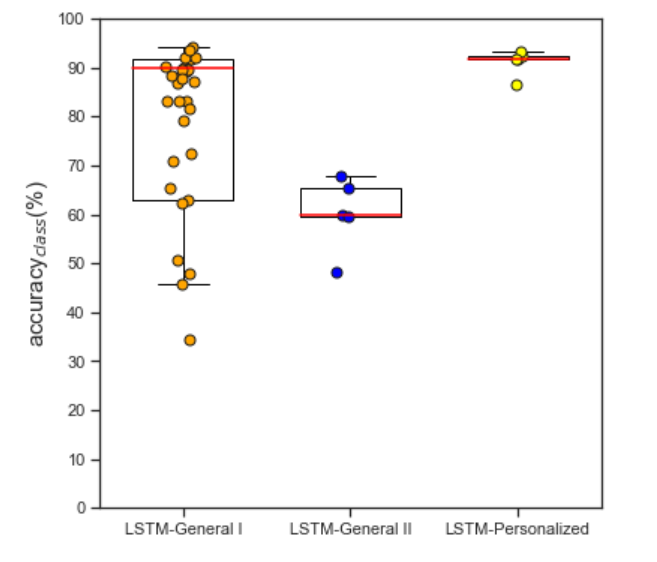}
\includegraphics[width=6cm]{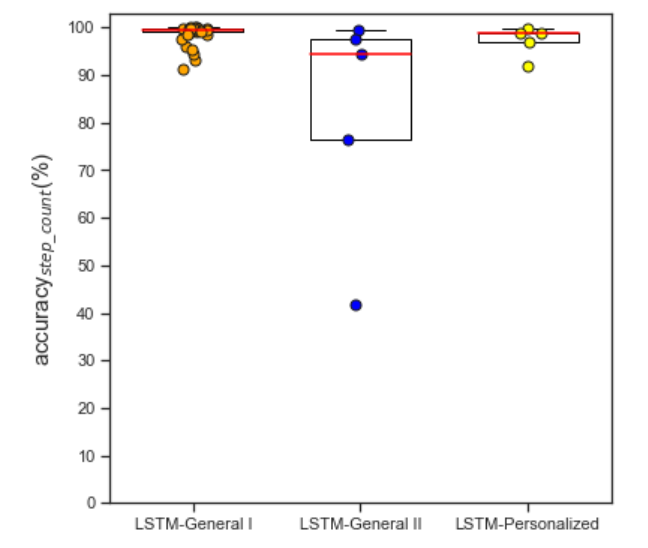}\\
\hspace{0.6cm}(a) \hspace{5.7cm}(b)\\
\caption{Comparison of experiments at subject level: (\textbf{a}) $accuracy_{class}$ box plot. (\textbf{b}) $accuracy_{steps}$ box plot.}.
\end{figure} 
From the boxplots shown in Fig. 1., drastic decreases in performance of LSTM-General II in case of outliers is also adequately handled by LSTM-Personalized. Additionally, the median performance of LSTM-Personalized and LSTM-General I are similar in terms of both the metrics, which suggests information from different devices and different subjects can be handled by using personalized models.

\section{Conclusion}
In this paper, we proposed a personalized step counting algorithm built using an LSTM network and tuned with subject-specific data through domain adaptation. This shows that data collected from different subjects and devices can be modelled using our algorithm. Additionally, results suggest very little labelled data is necessary to construct personalized models. This work suggests deep transfer learning has significant potential to achieve clinically acceptable accuracy in the assessment of physical activity for multiple disease scenarios. Moreover, accurately identifying class labels is important as it helps in tasks like distance estimation, indoor localization, and human activity recognition. Therefore, in the future, we plan to conduct a large-scale study in patient populations to evaluate other wearable sensor related tasks, disease endpoints, and unsupervised domain adaptation.

%
%
%

\begin{thebibliography}{8}
\bibitem{tudor-locke}
Tudor-Locke C: Steps to Better Cardiovascular Health: How Many Steps Does It Take to Achieve Good Health and How Confident Are We in This Number? Curr Cardiovasc Risk Rep. \textbf{4}(4) (2010)

\bibitem{schmidt}
Schmidt MD, Cleland VJ, Shaw K, Dwyer T, Venn AJ: Cardiometabolic risk in younger and older adults across an index of ambulatory activity. Am J Prev Med. \textbf{37}(4) 278--284 (2019)

\bibitem{kraus}
Kraus WE, Yates T, Tuomilehto J, et al: Relationship between baseline physical activity assessed by pedometer count and new-onset diabetes in the NAVIGATOR trial. BMJ Open Diabetes Research and Care. \textbf{6}e000523 (2018)

\bibitem{edel}
M. Edel and E. Köppe: An advanced method for pedestrian dead reckoning using BLSTM-RNNs. International Conference on Indoor Positioning and Indoor Navigation (IPIN). 1--6 (2015)

\bibitem{chen}
Ziyi C: An LSTM Recurrent Network for Step Counting. arXiv preprint arXiv:1802.03486 (2018)

\bibitem{ryan}
R. Mattfeld, E. Jesch and A. Hoover: A new dataset for evaluating pedometer performance. 2017 IEEE International Conference on Bioinformatics and Biomedicine (BIBM), Kansas City, MO. 865--869 (2017)

\bibitem{pachi}
Pachi, A. \& Ji, Tianjian.s: Frequency and velocity of people walking. Structural Engineer \textbf{83}, 36--40 (2015)

\bibitem{avi}
Bleiweiss, A.: LSTM Neural Networks for Transfer Learning in Online Moderation of Abuse Context. ICAART \textbf{2}, 112-122 (2019)

\bibitem{feng}
Yuanyuan F et al.: Comparison of tri-axial accelerometers step-count accuracy in slow walking condition. Gait \& Posture. 11--16 (2017)
\end{thebibliography}
%

\end{document}